\documentclass[11pt,a4paper]{article}
\usepackage{EMNLP2022}
\usepackage{times}
\usepackage{array}
\usepackage{latexsym}
\usepackage{xcolor}
\usepackage{adjustbox}
\usepackage{multirow}
\usepackage{amsmath,amsfonts,amsthm,bm}
\usepackage{amssymb}
\usepackage{textgreek}
\usepackage{algorithm}
\usepackage{algpseudocode}
\usepackage{graphicx}
\usepackage{multicol}
\usepackage{lipsum}
\graphicspath{ {./images/} }

\usepackage[T1]{fontenc}

\usepackage[utf8]{inputenc}

\usepackage{microtype}

\usepackage{inconsolata}

\title{Readability Controllable Biomedical Document Summarization}
\author{Zheheng Luo \and Qianqian Xie\thanks{\phantom{h}Corresponding author} \and Sophia Ananiadou \\
         NaCTeM, The University of Manchester \\ \{zheheng.luo, qianqian.xie, sophia.ananiadou\}@manchester.ac.uk}

\date{}

\DeclareUnicodeCharacter{2212}{-}
\DeclareUnicodeCharacter{3002}{-}

\begin{document}
\maketitle
\begin{abstract}
Different from general documents, it is recognised that 
the ease with which people can understand a biomedical text is eminently varied, owing to the highly technical nature of biomedical documents and the variance of readers' domain knowledge. 
However, existing biomedical document summarization systems have paid little attention to readability control, leaving users with summaries that are incompatible with their levels of expertise.
In recognition of this urgent demand, we introduce a new task of readability controllable summarization for biomedical documents, which aims to recognise users' readability demands and generate summaries that better suit their needs: technical summaries for experts and plain language summaries (PLS) for laypeople.
To establish this task, we construct a corpus consisting of biomedical papers with technical summaries and PLSs written by the authors, and benchmark multiple advanced controllable abstractive and extractive summarization models based on pre-trained language models (PLMs) with prevalent controlling and generation techniques.
Moreover, we propose a novel masked language model (MLM) based metric and its variant to effectively evaluate the readability discrepancy between lay and technical summaries.
Experimental results from automated and human evaluations show that though current control techniques allow for a certain degree of readability adjustment during generation,
the performance of existing controllable summarization methods is far from desirable in this task.

\end{abstract}

\section{Introduction}
Automatic summarization for biomedical documents~\cite{guo2020automated,deyoung2021ms2} such as clinical literature~\cite{wang2020cord,deyoung2021ms2}, provides an efficient way for readers to acquire desirable biomedical information quickly.
Unlike general documents, biomedical documents have characteristics of mounting scientific jargon~\cite{10.7554/eLife.27725}, and complex language structures~\cite{friedman2002two}.
Therefore, readers such as non-experts and professionals would seek textual information on different readability levels, since the variance of their biomedical knowledge affects their ease of understanding biomedical papers. 
For example, an in-domain expert might require accurate and clear technical summaries with medical jargon and professional language, to quickly grasp the main contributions of biomedical papers. In contrast, layperson readers usually require plain language summaries with less technical terms and more context of the research, which are easier to understand.
\begin{figure}[t]
  \includegraphics[width=0.45\textwidth]{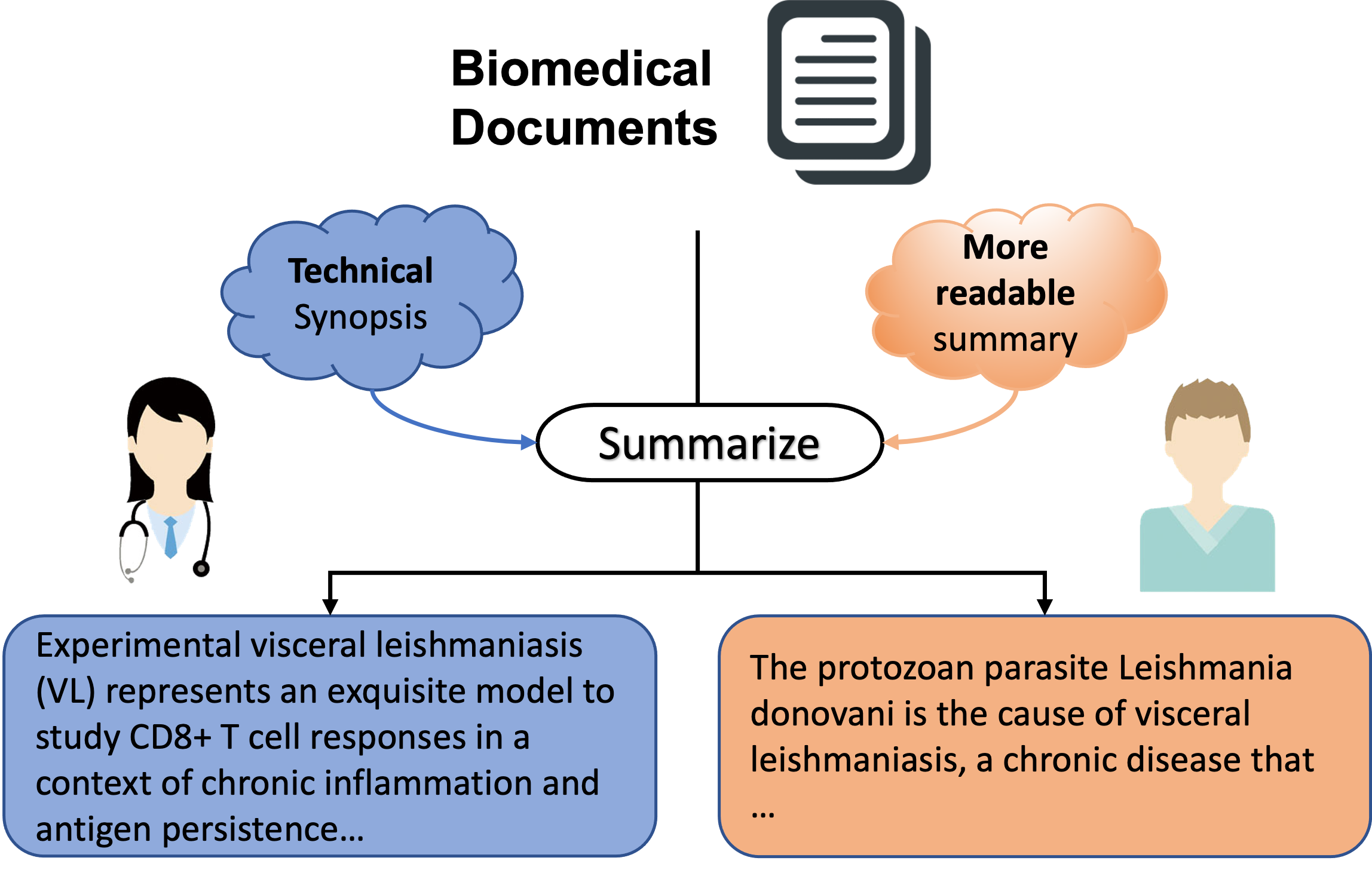}
  \caption{Example of our task. Summaries are generated according to users' demand for readability.}
  \label{fig:1}
\end{figure}
Nevertheless, current biomedical summarization systems are only able to offer technical abstracts~\cite{sotudeh2020attend,deyoung2021ms2,xie2022gretel,xie2022pre,bishop2022gencomparesum} or lay language summaries~\cite{guo2020automated,chandrasekaran-etal-2020-overview-insights}, fail to generate compatible summaries for various users according to their levels of expertise without considering the readability as an aspect to be controlled during summary generation~\cite{he2020ctrlsum}.
We argue that it is urgent to develop biomedical summarization approaches that can not only condense biomedical documents into concise summaries but also adjust the readability level of summaries to improve the dissemination of scientific information.

Our research aims to tackle the problem, and thus propose a novel task of readability controllable biomedical document summarization, 
which is to automatically recognize users' readability demands and generate summaries that are compatible with their expertise level and needs, as shown in Figure \ref{fig:1}. 
Specifically, in a binary readability level controlling setting, it is to produce technical summaries for experts, while plain language summaries (PLS) for laypeople.
The task is challenging since: 1) it requires the model to accurately recognize different readability demands from limited guiding signals,
2) it requires a suitable selection of content from long biomedical documents for various readers guided by their readability demands,
3) it requires the model to learn not only lexical and syntactic adjustment but also paraphrasing according to users' needs.  Since professionals pay more attention to clarity and accuracy while non-experts prefer summaries that are easier to understand.

To approach this task, we build the first corpus consisting of 28,124 biomedical literature with technical and plain language summaries written by the authors, then conduct a thorough analysis of the collected data including statistics, readability metrics, and textual features. Next, we examine several controlling techniques on prevalent pre-trained language models (PLMs) and evaluate their performance on our dataset. Apart from automatic assessment, we also bring in  the human evaluation due to the inefficacy of current metrics for readability and text generation.
To better characterize readability differences between technical summary and PLS, we further propose a novel masked noun phrase-based text complexity metric and its variant based on the masked language model (MLM).
It is superior in modelling the semantic structure of biomedical texts compared to traditional metrics and existing MLM-based metrics.

Overall, our main contributions are summarised as follows: (1) We introduce a novel task of readability controllable biomedical document summarization. (2) We build a corpus\footnote{can be downloaded from \url{http://www.nactem.ac.uk/readability/}} with 28,124 biomedical papers with their technical and plain language summaries, which will facilitate further exploration in this task. (3) We propose an MLM-based text complexity metric, which surpasses existing readability evaluation metrics on our dataset. (4) We examined controlling techniques including prompts and multi-heads on both extractive and abstractive methods to adjust readability during summarization and found the performance is far from satisfying. To the best of our knowledge, this is the first effort to consider readability as a controllable attribute in scientific document summarization.

\section{Related Work}
\subsection{Biomedical Text Summarization}
Neural networks and PLMs have been explored for biomedical document summarization in recent years, due to their success in general text summarization~\cite{cohan2018discourse,DBLP:conf/emnlp/LiuL19,DBLP:journals/corr/abs-1902-09243,wang2021pre}.
\citet{sotudeh2020attend} improved radiology report summarization by incorporating medical ontology into a sequence-to-sequence summarizer.
\citet{DBLP:journals/corr/abs-2008-11293} investigated the BART model~\cite{DBLP:conf/acl/LewisLGGMLSZ20} with domain specific pre-training strategies and input decorations for multi-document summarization of randomized controlled trials (RCTs).
Progress in biomedical summarization has also been advanced due to the emergence of in-domain corpora. \citet{cohan2018discourse} and \citet{wang2020cord} compiled a large amount of biomedical literature with their abstracts as summaries. \citet{deyoung2021ms2} investigated if systematic reviews could be summarised from their cited clinical trials.
\citet{guo2020automated} mixed summarization and simplification by generating plain language summary conditioned on abstracts of systematic reviews. 

\subsection{Controllable Text Summarization}
\label{subsec:control}
Recent efforts on controllable text summarization mostly focus on news articles. \citet{fan2018controllable} has leveraged PLMs with special tokens prepended to the input, to control the length, entities, and style of the generated summary. 
\citet{DBLP:conf/coling/ZhengCZL20} and \citet{he2020ctrlsum} further extended prompts, keywords and entities as guiding markers.
\citet{chan2021controllable} proposed the constrained markov decision process (CMDP) based method to control attributes of summarization.
Other works have tried exerting control in decoding.
HydraSum~\cite{goyal2021hydrasum} distributed different values of an attribute into multiple decoders and leveraged a gate mechanism to gain control over properties such as abstractness and length.
\citet{DBLP:conf/emnlp/AmplayoAL21} and \citet{DBLP:conf/eacl/AmplayoL21} focused on the aspect control of opinion summarization on reviews.
To the best of our knowledge, our work is the first effort to consider readability as a controllable attribute in scientific document summarization, which is important for specific-domain, especially biomedical science.

\subsection{Readability Metrics}
Readability is defined as the ease with which a reader can understand a piece of text. Many factors are involved in determining readability, such as lexical and syntactic sophistication, discourse cohesion, and background knowledge~\cite{crossley2017predicting}.
Prior work on lay summarization~\cite{guo2020automated} evaluated their corpus by traditional readability formulas like Flesch-Kincaid Grade Level~\cite{kincaid1975derivation} which is inefficient in revealing the readability differences in scientific writings. \citet{Martinc2021} has shown the potential of the PLM in estimating text readability. \citet{devaraj2021paragraph} used an MLM-based metric to better classify technical abstracts and PLS of medical reviews.
In this work, we propose an advanced MLM-based metric to manifest the readability differences among summaries in our corpus and evaluate the output of tested models. 

\section{Task Overview}
\noindent\textbf{Definition.} The objective of this task is to generate summaries of biomedical documents on different readability levels based on users' demands.
Let $D=\{d_1,d_2,\cdots,d_k\}$ denotes the set of source documents, each document $d_i = \{x_{i,1},x_{i,2},\cdots,x_{i,n}\}$ can be represented by the sequence of $n$ tokens, $S_i$ stands for the target summary of document $d_i$ that is represented by the sequence of $m$ tokens:$\{s_{i,1}, s_{i,2},\cdots,s_{i,m}\}, m \ll n$. 
$r$ means the readability level the user might want. 
The task can be formulated as a conditional generative problem as follows:
\begin{equation} \label{eq1}
\small
P(S|D, r) = \prod_{i}^{k}P(S_i|d_i,r)
\end{equation}
which maximizes the probability of generating \textit{S}  when given the document set \textit{D} and the readability demand \textit{r}. 
In this work, since the exploration of readability controlling summarization is still in an initial stage, we start with single document input with a binary readability control between "technical" and "plain language" and leave more fine-grained control to future work. 
We consider $r_t$ mean the demand for \textit{technical summary} that is suitable for experts, while $r_p$ means the demand for \textit{plain language summary} (PLS) for laypeople readers.
Thus, we have both technical target summary $S_i^t$ and plain language target summary
$S_i^p$ for each input document $d_i$, to train the model. Additionally, a technical summary and a PLS generated from the same document by the same model will be referred to as a pair of summaries in this paper.


\noindent\textbf{Evaluation.} The most commonly used metric for evaluating summarization models is ROUGE~\cite{lin2004rouge}, which has served as a standard in various text generation tasks. However, a recent study~\cite{bhandari2020re} has shown that ROUGE scores do not always agree with human evaluation when assessing generated summaries. Also, traditional readability metrics are found unable to show the significant readability difference between the technical summary and their simplified counterparts~\cite{devaraj2021paragraph}. 
Thus, we conducted both automatic and human evaluations to assess the readability and general qualities of generated summaries.
\begin{table}
\centering
\small
\begin{tabular}{p{0.115\textwidth}p{0.06\textwidth}m{0.055\textwidth}m{0.055\textwidth}m{0.055\textwidth}}
\hline \textbf{Dataset} & \textbf{docs} & \textbf{avg. doc length} & \textbf{avg. abs length} & \textbf{avg. PLS length}\\ \hline
PubMed & 133,000  & 3,016 & 203 & -\\
CDSR & 7,805 & -& 714 &374\\
Ours & 28,124 & 6,697 & 287 & 204 \\ \hline
\end{tabular}
\caption{Statistics of our PLOS datasets compared with existing biomedical summarization corpora PubMed~\cite{cohan2018discourse} and CDSR~\cite{guo2020automated}}
\label{table1}
\end{table}

\section{Dataset Description}
\label{sec dataset description}
\subsection{Data Compilation}
We constructed the corpus consisting of peer-reviewed biomedical research papers with the technical summaries and PLSs from journals including PLOS \footnote{\url{https://journals.plos.org/plosone/}} Biology, PLOS Computational Biology, PLOS Genetics, PLOS Medicine,
PLOS Neglected Tropical Diseases, and PLOS Pathogens, cover a broad range of biomedical research subjects. 
The PLSs are placed under the section \textit{Author Summary} in the format of the PLOS articles and written by the authors following the requirement of PLOS submission guidelines which suggest highlighting where the work fits within a broader context, presenting the significance simply and avoiding the use of acronyms and complex terminology \footnote{\url{https://journals.plos.org/ploscompbiol/s/submission-guidelines}}.

To build the corpus, we downloaded the whole PLOS article dataset (up to 4th April 2022) \footnote{\url{https://plos.org/text-and-data-mining/}}. Then, after filtering out papers without plain language summaries, we extracted the full text, the abstract as technical summary, and the \textit{Author Summary} as PLS from the remaining papers, resulting in 28,124 document-technical summary-PLS triplets. We randomly sample 1,000 triplets respectively for validation and test, leaving the rest 26,124 for training. 
Table \ref{table1} shows the main statistics of our dataset and other biomedical summarization corpora. 
Compared to previous work, our dataset is the first that contains both technical summaries and PLS, and our source documents are much longer, making the task more challenging.

\subsection{Quantitative Analysis} 
\label{subsec:qa}
To investigate the readability differences between technical summaries and PLSs, we examined various metrics and calculated Spearman's correlations with the readability levels of summaries in our dataset.

\noindent\textbf{Traditional readability formula.} Three established heuristics-based readability metrics are adopted, including the Flesch-Kincaid Grade~\cite{kincaid1975derivation}, Coleman-Liau Index~\cite{coleman1975computer}, and automated readability index (ARI)~\cite{senter1967automated}, which are used to approximate the U.S. grade level to understand a written text. 
These metrics rely on shallow features like the length of a sentence or the number of characters in words and thus are unable to fully reveal the gap between the summaries of biomedical documents on different readability levels~\cite{gao-etal-2019-automated}.

\noindent\textbf{Language model based metric.} Predicted probabilities for masked tokens by language models have shown effectivity in measuring readability. \citet{devaraj2021paragraph} assumed that language models trained on general domain text would align lower likelihoods to tokens in technical jargon and higher likelihoods to common tokens. 
They proposed the BERT~\cite{DBLP:conf/naacl/DevlinCLT19} and SciBERT~\cite{DBLP:conf/emnlp/BeltagyLC19} based metric, which randomly masks $15\%$ of tokens in each sentence, and computes the average likelihoods of original masked tokens in the distributions from the model output.
This method outperforms traditional metrics such as Flesch-Kinard Grade in distinguishing plain language summaries against technical abstracts of systematic reviews.
We will refer to their metric as masked random token-based text complexity (MRTTC).

\noindent\textbf{Masked noun phrase-based metric.} It is recognized there are many technical terms existing in biomedical texts in the form of noun phrase (NP), which should be considered as a complete semantic unit. Thus token-level masking has its limitation since random token-level masking may corrupt the semantic integrity of technical terms in biomedical texts.
Rather than randomly masked tokens, the likelihood of whole noun phrases predicted by a language model could be a finer indicator to discriminate between plain language and technical texts. 
Therefore, we propose the novel masked noun phrase-based text complexity metric, denoted as NPTC.
We first leverage Spacy~\cite{spacy2} to extract all NPs in each document, then filter out NPs that only consist of stop-words to prevent disturbance of common tokens. 
Next, we mask tokens of each NP in the summaries of each document, to create a masked summary and use a BERT pre-trained on general text \footnote{\url{https://huggingface.co/bert-base-uncased}} to predict the probability sequence of masked NPs. Lastly, the likelihoods of the target tokens in each masked NP are averaged as the likelihood of the NP, which is further averaged across the document to achieve its final score. 

Moreover, we follow~\citet{martinc2021supervised} who argue that compared to directly sum word negative log-likelihood (WNLL), assigning weights to WNLL depending on the ranking of their values in a text is a more effective way to model readability given that several unreadable words might drastically increase the difficulty of the entire text. Based on these considerations, we propose the ranked NP-based text complexity (RNPTC). Specifically, after obtaining the probability of each NP, we sort them in descending order, and then use the following formula to average over the probabilities:
\begin{equation}
\small
RNPTC = - \log(\frac{\sum^{|NPs|}_{i=1} P_{NP(i)} / \sqrt{i}}{|NPs|})
\label{eq2}
\end{equation}
where $|NPs|$ denotes the total amount of NPs, i stands for the rank of the current NP, and $P_{NP(i)}$ is the probability of the NP that is ranked $i$. 
By assigning the reciprocal of square rooted ranking as the weight to every NP, the complexities of the most difficult NPs are emphasised and the interference of common phrases is reduced.
The detailed process of RNPTC is illustrated in Algorithm \ref{alg:cap} at Appendix \ref{sec:al}. 
For a clearer comparison, we take the negative logarithm of ~\citet{devaraj2021paragraph}'s metric in the experiment.

To illustrate the effectivity of our proposed metrics, we compare Spearman's correlations of examined metrics with the readability levels on our dataset and the CDSR dataset(Cochrane Database of Systematic Reviews) of which the detailed collecting process is illustrated in Appendix \ref{sec append cdsr clt}. The results are shown in Table \ref{table-plos} and \ref{table-csdr}. 
Compared to PLM-based metrics including RNPTC, NPTC and MRTTC, traditional formulas (C-L Index, F-K Grade, ARI) have lower Spearman's correlation scores.
This indicates that they are not good indicators of readability differences between technical texts and plain language ones due to their dependence on shallow statistical features.
Our proposed metrics surpasses all other methods in correlation with the readability level in both datasets. 
This shows that the consideration for fine-grained semantic structure of biomedical texts in our metrics RNPTC and NPTC is helpful to discriminate readability differences between technical summaries and PLS. 

\begin{table}
\centering
\footnotesize
\begin{tabular}{p{0.12\textwidth}m{0.075\textwidth}m{0.075\textwidth}m{0.10\textwidth}}
\hline \textbf{Metric} & \textbf{Avg. on TS} & \textbf{Avg. on PLS} & \textbf{Spearman's \textrho}\\ \hline
RNPTC & 2.63 & 2.45 & 0.337 \\
NPTC & 1.84 & 1.72 & 0.183\\
MRTTC& 0.81 & 0.78 & 0.148 \\
C-L Index & 16.50 & 15.93 & 0.140\\
F-K Grade & 14.99  & 14.76 & 0.065\\
ARI & 17.90 & 17.41 & 0.102\\

\end{tabular}
\caption{Average values of different metrics in our dataset and their Spearman's correlation with readability. TS means technical summaries.}
\label{table-plos}
\end{table}

\begin{table}
\centering
\footnotesize
\begin{tabular}{p{0.12\textwidth}m{0.075\textwidth}m{0.075\textwidth}m{0.10\textwidth}}
\hline \textbf{Metric} & \textbf{Avg. on Abstract} & \textbf{Avg. on PLS} & \textbf{Spearman's \textrho}\\ \hline
RNPTC & 2.78 & 2.46 & 0.584 \\
NPTC & 1.81 & 1.50 & 0.500\\
MRTTC& 0.93 & 0.81 & 0.513 \\
C-L Index & 14.71 & 14.22 & 0.138\\
F-K Grade & 11.80  & 12.67 & -0.190\\
ARI & 14.24 & 15.43 & -0.217\\

\end{tabular}
\caption{\label{font-table} Average values of different metrics in the abstracts and PLS of CSDR and their Spearman's correlation with readability.}
\label{table-csdr}
\end{table}
\begin{table*}[t]
\centering
\scriptsize
\begin{tabular}{>{\raggedright\arraybackslash}m{0.1\textwidth}>{\centering\arraybackslash}m{0.1\textwidth}>{\raggedright\arraybackslash}m{0.34\textwidth}>{\raggedright\arraybackslash}m{0.34\textwidth}}\\\hline 
& \centering Distinction & Technical Summary & Plain Language Summary \\\hline
\multirow{2}*{Content} & \shortstack{Experiment \\Details} &
1300 individuals with minor TT were recruited and randomly assigned to ... \textcolor{magenta}{At two-years all epilation arm participants were offered free surgery. At four-years 1151 (88.5\%) were re-examined: 572 (88\%) and 579 (89\%) from epilation and surgery arms, respectively. At that time, 21.1\% of the surgery arm participants had recurrent TT; 189/572 (33\%) of the epilation arm had received surgery, } 
& We conducted a randomized controlled trial of epilation versus eyelid surgery (the main treatment option) in 1300 people with mild trichiasis in Ethiopia.\\\cline{2-4}
& \shortstack{Background \\Information} &  
West Nile virus (WNV) is transmitted to vertebrate hosts by mosquitoes as they take a blood meal. & \textcolor{cyan}{Since it was first introduced into the United States in 1999, West Nile virus (WNV) has caused significant disease in humans, horses, and other animals.} WNV is transmitted to humans and other ... \\\hline
\multirow{2} * {Language} & \shortstack{Scientific \\Jargon} &
\textcolor{purple}{Experimental visceral leishmaniasis (VL)} represents an exquisite model to study CD8+ T cell responses in a context of \textcolor{purple}{chronic inflammation and antigen persistence}, since it is characterized by \textcolor{purple}{chronic infection in the spleen} and CD8+ T cells are required for the \textcolor{purple}{development of protective immunity.} 
& Preliminary studies suggested that CD8+ T cells can \textcolor{green}{kill parasites and limit disease}; however, studying these important \textcolor{green}{killer cells} has been \textcolor{green}{hindered}, because we do not know what parasite molecules they recognize\\\cline{2-4}
& \shortstack{Syntactic \\structure} & Hookworm-related cutaneous larva migrans (CLM) is a common but neglected tropical \textcolor{pink}{skin disease caused by} the migration of animal hookworm larvae in the epidermis. & Hookworm-related cutaneous larva migrans (CLM) is a parasitic \textcolor{teal}{skin disease} common in developing countries with hot climates. \break \textcolor{teal}{The disease} is caused by animal hookworm larvae...\\\hline
\end{tabular}
\caption{Difference between the writing styles of technical summary and PLS. Red texts are contents which are included in technical summaries and omitted in PLS. Blue texts indicate complimentary information provided for PLS but not for technical summaries. Green texts represent more layperson-friendly expression while purple means scientific jargon. Teal texts stand for simplified syntactic structures of texts in pink.}
\label{table3}
\end{table*}

\subsection{Qualitative Analysis} To better demonstrate how the authors adjust the readability when writing summaries for different audiences, we manually read through some random samples from our dataset, and suggest the distinctions of the writing styles are rooted in their choice of content and language. Under these two major subjects, we further defined four specific aspects and present them in Table \ref{table3}.

\noindent\textbf{Content.} When choosing content for writing summaries on different readability levels for their work, the authors usually keep the general compositions similar. But they are likely to include detailed experimental design and quantitative data which are helpful in showing the confidence of the experiments when drafting a technical summary while omitting these descriptions in PLS. Moreover, we have noticed that authors usually add a few sentences at the beginning of PLS to explain the subject of their research and the context around it, smoothing laypeople readers' understanding of the research question.

\noindent\textbf{Language.} From the perspective of language use, there are two main distinctions. First, scientific jargon has been heavily used in technical summaries for accuracy and conciseness, but either removed or replaced with more common expressions in PLS. Second, the authors are also observed to use simpler syntactic structure in PLS to enhance readability.

\section{Baselines for Readability Controllable Summarization}
Existing text summarization methods can be classified into two main ways, extractive and abstractive approaches. 
We carry out experiments with both methods in this task.

\subsection{Skeleton Model}
Prevalent PLMs such as BERT~\cite{DBLP:conf/naacl/DevlinCLT19} and BART~\cite{lewis2019bart} using full attention mechanisms are faced with the out-of-memory problem when training with long sequences. Nevertheless, the average amount of words in a biomedical scientific document is usually several thousand, and shortening the input into hundreds of tokens would significantly reduce information that could be used by summarization models. Thus we take on Longformer-Encoder-Decoder (LED)~\cite{beltagy2020longformer} as the underlying architecture. LED is implemented with local attention based on a sliding window for all tokens in the sequence and global attention for certain tokens to learn task-specific representation. By the double attention design, LED reduces the computing requirements for long sequences and has shown better performance in summarizing long documents like Pubmed and Arxiv papers. 

\subsection{Control Methods}
\noindent\textbf{Control Prompts.} Prior works~\cite{he2020ctrlsum} aforementioned in Section \ref{subsec:control}  have proved that combined with PLMs, prompts like special tokens are effective to guide the model to generate controlled text while training on corresponding data with certain attributes. Thus we design two types of markers \(|PLAIN|\) and \(|PRO|\) to respectively represent signals for producing PLS and technical summaries. During training, the special tokens are prepended to the source text to form the input source documents for their corresponding summaries. 
Also under the double attention setting in LED, to fully exert the guiding effect of these special tokens, we set them to conduct global attention with other tokens. In abstractive mode, we train the LED in a sequence-to-sequence style and used beam search during inference. For the extractive version, we followed the design in \citet{liu2019text} by taking the encoder of LED and putting two transformer layers with a final classifier upon it. Then an end-of-sentence token is inserted after every sentence in the source documents and their hidden states output from the LED encoder are used as representations of corresponding sentences into the following extractive layers. The model is optimized to pick sentences that appear in the summaries extracted by a greedy selection method \cite{liu2019text} which will be referred to as Oracle extraction hereinafter. During inference, we select top-\textit{k} sentences by their scores as the final summary.

\noindent\textbf{Multi Heads.} \citet{goyal2021hydrasum} has shown that the attributes of generated summaries like length can be distributed into multiple decoders in a sequence-to-sequence structure, then by decoding from different heads the user can obtain either long or short summaries. Inspired by their multi heads design, we tried adding one more decoder into the LED model for generating summaries of different readabilities. Specifically, in our abstractive framework, the encoder is kept shared but two decoders are set independent of learning either technical or plain language style generation. We trained this multi-decoder model with the gate mechanism in which the probabilities predicted by the heads of two decoders would be multiplied by $1-g$ and $g$ respectively.  We adjust $g$ equals 1 for technical summaries generation and 0 for PLS. During inference, we  sample summaries from the two heads separately. In the extractive multi-heads, we keep the encoder shared and create two extractive heads to select sentences for different readabilities. The model is trained via the same gate mechanism as the abstractive one. In both models, we set the start of sentence token \(\langle s\rangle\) in LED for global attention.
Due to length limitations, please see Appendix \ref{sec:es} for detailed experimental setups.
%

\section{Results}
\subsection{Automatic Evaluation}
In Table \ref{table4}, we list the ROUGE scores of generated summaries from five tested methods.
\begin{table}
\centering
\footnotesize
\begin{tabular}{ccccc}
\hline Method &  & R1 & R2 & RL \\ \hline
\multirow{2}{*}{Oracle} & tech & 62.79 & 31.36 & 58.26  \\
& plain & 57.08 & 25.31 & 52.39\\\cline{2-5}
\multirow{2}{*}{CP Abs}& tech & 48.51 & 16.89 & 44.53\\
& plain & 45.69 & 15.40 & 41.89 \\\cline{2-5}
\multirow{2}{*}{CP Ext}& tech & 49.52&16.63&	45.50\\
& plain &45.30&13.00	&40.99 \\\cline{2-5}
\multirow{2}{*}{MH Abs}& tech&  \textbf{49.87}&\textbf{ 17.85} & \textbf{46.02} \\
& plain&\textbf{47.31}  & \textbf{14.73}& \textbf{43.08} \\\cline{2-5}
\multirow{2}{*}{MH Ext}& tech& 49.17 & 16.75 &39.59  \\
& plain&45.40  &13.03  &40.99  \\ \hline
\end{tabular}
\caption{ROUGE scores of generated summaries, CP stands for Control Prompts and MH is the abbreviation of Multi Heads}
\label{table4}
\end{table}
The upper bound of ROUGE score is established by the Oracle extraction in the first 2 rows, which displays a distinct lexical difference between technical summaries and plain language summaries since the lower ROUGE scores indicate that PLS are harder to approach by directly selecting sentences from the document.
For the same reason, though competitive in summarizing technical summaries, extractive methods are largely outperformed by abstractive methods in generating PLS, manifesting the advantage of abstractive methods in adjusting the style of expression.
Among all methods, multi heads abstractive LED (MH Abs) performs best on ROUGE of both kinds of readability.
This can be attributed to the additional parameters provided by the multi heads technique for a model to adapt to generation for different readability demands.
However, there is no evident difference between the two control techniques on extractive methods, indicating that under our training setting, more parameters are not helpful for controlling in an extractive way.

In Table \ref{table5}, we further show the difference in readability scores between pairs of summaries evaluated by RNPTC and three traditional functions. The values of target summaries in the test set (PLOS Test) are added in the last row for comparison.
\begin{table}[t]
\centering
\footnotesize
\begin{tabular}{cccccc}
\hline Method && FKG & CLI & ARI & RNPTC\\ \hline
\multirow{2}{*}{Oracle} & tech &14.37&14.79&	17.02&2.57 \\&plain&14.46&15.01&17.00&2.50\\\cline{2-6}
\multirow{2}{*}{CP Abs}& tech&14.85	&15.59&	17.40	&2.46\\
&plain &15.13&	15.83&	17.73&	2.40\\\cline{2-6}
\multirow{2}{*}{CP Ext}& tech&14.91&15.60&17.43	&2.60\\
&plain &15.04&15.80&17.58&2.53\\\cline{2-6}
				
\multirow{2}{*}{MH Abs}& tech&14.25	&15.07&	16.63&2.41\\
&plain &14.75	&15.12	&17.18&2.35\\\cline{2-6}
\multirow{2}{*}{MH Ext}& tech&15.02	&15.65&17.57	&2.60\\
&plain &15.13	&15.91	&17.73&2.51\\\cline{2-6}
\multirow{2}{*}{PLOS Test}& tech&14.90	&16.45&	17.75&2.57\\
&plain &14.80&	15.98&	17.43&2.41\\ \hline
\end{tabular}
\caption{Readability scores of generated summaries}
\label{table5}
\end{table}
For technical summaries, the generated summaries should have higher readability scores which indicate more complicated tokens such as terms that could embed key information, while output for PLSs is expected to get lower scores by using less jargon but more common words. 
From traditional readability scores, most generated PLSs are of no difference or even harder to read than their technical counterparts. 
As mentioned before in section \ref{subsec:qa}, these traditional readability formulas can largely be influenced by shallow statistic features such as the number of tokens and sentences.
Thus, they may be deficient to reflect the readability differences. 

Unlike these metrics, our proposed RNPTC demonstrates that the text complexity is generally lower in PLSs.
However, the variance is slight compared to that between pairs of target summaries, which indicates these controlling models are not ideal to adjust their output under different readability demands.
Moreover, from RNPTC, PLSs and technical summaries generated via extractive methods are generally more complicated than those from abstractive methods and target summaries, suggesting the high lexical complexity of sentences from original documents.
With regards to generated summaries of abstractive methods, their RNPTCs are lower than those of target summaries especially for technical ones, implying they might fail to provide enough key information for expert readers.  
This could be caused by the lower contextual tendency of PLM-based text generation
~\cite{gehrmann2019gltr}.
Thus when generating summaries, these PLM-based models tend to select tokens that will maximize the global probability while avoiding technical terms since their predicting probabilities can be lower than common tokens.

\subsection{Textual Variance}
Observing the slight difference in numerical readability metrics between pairs of generated summaries, we wonder how much textual variance can controlling techniques actually lead to when summarizing for different readability demands.
Hence, we take the ratio of the number of n-grams in a PLS that appear in the corresponding technical summary to the total number of n-grams in the PLS as the indicator for similarity. It is assumed that fewer overlapping suggests stronger control of a summarization model. In Table \ref{table6}, we compared the n-gram overlapping fractions among pairs of target summaries in our whole PLOS dataset(PLOS Whole) and the test subset(PLOS Test) as well as pairs of generated summaries from the five examined methods.

Between pairs of target summaries, the similitude is evidently smaller compared to all generated results, showing the limited ability of the controlling techniques to adjust output according to readability demands.
Two extractive methods suffer the most from the overlapping problem, approximately 80 per cent of 4-grams in their PLSs are also in the technical summaries, which could be due to the already large n-grams overlapping fraction in the Oracle extraction. Meanwhile, training with more discernable pairs of target summaries, abstractive methods outperform extractive ones significantly in generating more distinctive contents when given different readability demands. However, the overlapping ratio is still high, matching the little difference in readability evaluation.

\begin{table}[htb]
\centering
\footnotesize
\begin{tabular}{ccccc}
\hline \textbf{Ngrams} & \textbf{N=1} & \textbf{N=2} & \textbf{N=3}& \textbf{N=4}\\ \hline
PLOS Whole& 0.67 & 0.25 & 0.12 & 0.08 \\
PLOS Test & 0.67 & 0.26 & 0.13 &0.08 \\
Oracle & 0.73 & 0.42 & 0.33 & 0.30 \\
CP Abs & 0.80 & 0.59 & 0.51 &0.47\\
CP Ext & 0.91 & 0.84 & 0.82 & 0.80\\
MH Ext&0.87  & 0.76 & 0.73& 0.72\\
MH Abs&\textbf{0.76}  & \textbf{0.47} & \textbf{0.35}& \textbf{0.29}\\
\end{tabular}
\caption{N-gram overlap fraction between PLSs and technical summaries}
\label{table6}
\end{table}

\subsection{Human Evaluation}
We further conduct a human evaluation to assess the two main aspects to be tested in our task: controlling performance and general quality. More details about the setup can be found in Appendix \ref{sec:appdx human}. Examples of evaluated target and generated summaries can be found in Appendix \ref{sec:appdx examples}, Table \ref{table: examples}.

When assessing controlling performance, we split evaluators into \textit{experts} and \textit{laypeople} according to their education background and ask them to rate the readability on a scale of 1-5 to see how biomedical expertise affects people's judgement on readability. In respect of general quality, we inform the evaluators to rate each generated summary from 1 to 5 in three aspects: 1) relevance: to what extent does the summary cover the important information \footnote{we here ask annotators to compare generated summaries to target summaries so the relevance of target summaries is omitted}. 2) grammar: how good the summary is grammatically. 3) coherence: is the content well-structured in the summary. The results are shown in Table \ref{tab:human}.
\begin{table}[t]
\centering
\footnotesize
\begin{tabular}{m{1.2cm}>{\raggedleft\arraybackslash}m{0.9cm}m{0.7cm}m{0.7cm}m{0.7cm}m{0.7cm}}
\hline \textbf{Aspect} &&\multicolumn{2}{c}{\textbf{Target}} & \multicolumn{2}{c}{\textbf{Generated}}\\ \cline{3-6}
&&\textbf{Tech} & \textbf{Plain} & \textbf{Tech} & \textbf{Plain}\\ \hline
\multirow{2}{*}{Readability} & \textit{experts} &4.90&4.83&4.50&4.45 \\
&\textit{laypeople}&3.37&4.43&3.83&3.90\\
Relevance&  &-	& -&3.39&3.54\\
Coherence& &4.62	&4.72	&4.01&4.05\\
Grammar&  &4.83&4.90&4.39&4.35\\\hline
\end{tabular}
\caption{Human evaluation of target and generated summaries}
\label{tab:human}
\end{table}

In the judgment of the \textit{experts}, the readability differences between pairs of target summaries are small while \textit{laypeople} evaluators evidently discern the gap between target technical summaries and PLS. In both groups, the scores of pairs of generated summaries are quite close. And for \textit{laypeople} group the readability difference between generated TS and PLS is modest compared to that between target technical summaries and PLSs, matching the slight effect of control shown in previous analyses. From the gap between the two groups' readability scores of technical summaries , we can see the influence of domain expertise on the ease of understanding biomedical texts.
In respect of general quality, we can find that grammatically generated texts are only slightly worse than human writings. However, when it comes to relevance, generated summaries achieve only mediocre performance, suggesting the difficulty of capturing key information in long biomedical documents. Also, the lower coherence of generated summaries reveals the inability of PLMs to keep the output sentences well-connected when generating paragraph-level texts.

\section{Conclusion}
In this work, we introduce a novel controlling summarization task that aims at encapsulating biomedical scientific literature on different readability levels. 
We draw the full text, abstracts, and plain language summaries from papers published in peer-reviewed biomedical journals to build the first corpus for the task.
We propose an effective MLM-based readability metric and its variant, which outperform traditional and previous MLM-based readability measures in distinguishing technical summaries against PLSs.
We leverage the Longformer-Encoder-Decoder (LED) as our skeleton model and examine prevalent generation controlling techniques on both extractive and abstractive methods. We found that though the abstractive approach combined with multi decoders can lead to higher-quality summaries and larger textual differences under binary readability demands compared to extractive methods, all tested models fail to show satisfying readability controlling ability. 
In the future, we would like to investigate how to introduce a more powerful teaching force such as language models trained with corpora of different readability levels to guide the controlling of summarization models. 
\section*{Limitations}
In this paper, we examined exerting control on the readability of the output of PLM-based summarization models.

Firstly, there are only texts of binary readability levels in the experimented corpus, limiting more fine-grained readability control under our supervised training methods.

Secondly, though distinctions of content and language are observed in pairs of generated summaries, the degree of control is still far from satisfying. We assume that introducing pre-trained discriminators~\cite{dathathri2019plug} into the summarization progress might help models push the output further towards a more technical or plain language direction.

Last but not least, due to the length and complexity of biomedical documents, we solely evaluate the relevance of generated summaries while leaving their faithfulness to source documents unchecked. Faithfulness or factuality of scientific summaries are of critical significance but receive little attention owing to the difficulty, we encourage future work to combine question generation and reading comprehension~\cite{wang2020asking} into the assessment of faithfulness in scientific document summarization.


\section*{Ethics Statement}
This paper presents a corpus built upon part of the whole article dataset from the PLOS corporation which is freely open to the public.
The advancement of large pre-trained language models has greatly boosted the improvement of summarization models in various domains including the biomedical area. High quality,  especially factual summaries would facilitate practice and research in both clinical and scientific communities. Yet, current state-of-the-art PLM-based models are unable to guarantee the faithfulness or factuality of generated summaries\cite{maynez2020faithfulness}. Thus we suggest any output of our proposed model should be manually examined by domain experts before using for any purpose.
\section*{Acknowledgement}
We would like to thank the evaluators (Jimin Huang, Feng Zhou, Shenhan Xie, Tairan Yang, Wang Yang, Yuandong Tao) for their valuable time and the Computational Shared Facility at The University of Manchester for offering computing devices.
We also thank Jimin Huang and Anonymous Reviewers for their suggestions and feedback on our paper.
\bibliographystyle{acl_natbib}
\bibliography{anthology,acl2021}

\begin{thebibliography}{43}
\expandafter\ifx\csname natexlab\endcsname\relax\def\natexlab#1{#1}\fi

\bibitem[{Amplayo et~al.(2021)Amplayo, Angelidis, and
  Lapata}]{DBLP:conf/emnlp/AmplayoAL21}
Reinald~Kim Amplayo, Stefanos Angelidis, and Mirella Lapata. 2021.
\newblock \href {https://doi.org/10.18653/v1/2021.emnlp-main.528}
  {Aspect-controllable opinion summarization}.
\newblock In \emph{Proceedings of the 2021 Conference on Empirical Methods in
  Natural Language Processing, {EMNLP} 2021, Virtual Event / Punta Cana,
  Dominican Republic, 7-11 November, 2021}, pages 6578--6593. Association for
  Computational Linguistics.

\bibitem[{Amplayo and Lapata(2021)}]{DBLP:conf/eacl/AmplayoL21}
Reinald~Kim Amplayo and Mirella Lapata. 2021.
\newblock \href {https://doi.org/10.18653/v1/2021.eacl-main.229} {Informative
  and controllable opinion summarization}.
\newblock In \emph{Proceedings of the 16th Conference of the European Chapter
  of the Association for Computational Linguistics: Main Volume, {EACL} 2021,
  Online, April 19 - 23, 2021}, pages 2662--2672. Association for Computational
  Linguistics.

\bibitem[{Beltagy et~al.(2019)Beltagy, Lo, and
  Cohan}]{DBLP:conf/emnlp/BeltagyLC19}
Iz~Beltagy, Kyle Lo, and Arman Cohan. 2019.
\newblock \href {https://doi.org/10.18653/v1/D19-1371} {Scibert: {A} pretrained
  language model for scientific text}.
\newblock In \emph{Proceedings of the 2019 Conference on Empirical Methods in
  Natural Language Processing and the 9th International Joint Conference on
  Natural Language Processing, {EMNLP-IJCNLP} 2019, Hong Kong, China, November
  3-7, 2019}, pages 3613--3618. Association for Computational Linguistics.

\bibitem[{Beltagy et~al.(2020)Beltagy, Peters, and
  Cohan}]{beltagy2020longformer}
Iz~Beltagy, Matthew~E Peters, and Arman Cohan. 2020.
\newblock Longformer: The long-document transformer.
\newblock \emph{arXiv preprint arXiv:2004.05150}.

\bibitem[{Bhandari et~al.(2020)Bhandari, Gour, Ashfaq, Liu, and
  Neubig}]{bhandari2020re}
Manik Bhandari, Pranav~Narayan Gour, Atabak Ashfaq, Pengfei Liu, and Graham
  Neubig. 2020.
\newblock Re-evaluating evaluation in text summarization.
\newblock In \emph{Proceedings of the 2020 Conference on Empirical Methods in
  Natural Language Processing (EMNLP)}, pages 9347--9359.

\bibitem[{Bishop et~al.(2022)Bishop, Xie, and
  Ananiadou}]{bishop2022gencomparesum}
Jennifer Bishop, Qianqian Xie, and Sophia Ananiadou. 2022.
\newblock Gencomparesum: a hybrid unsupervised summarization method using
  salience.
\newblock In \emph{Proceedings of the 21st Workshop on Biomedical Language
  Processing}, pages 220--240.

\bibitem[{Chan et~al.(2021)Chan, Wang, and King}]{chan2021controllable}
Hou~Pong Chan, Lu~Wang, and Irwin King. 2021.
\newblock Controllable summarization with constrained markov decision process.
\newblock \emph{Transactions of the Association for Computational Linguistics},
  9:1213--1232.

\bibitem[{Chandrasekaran et~al.(2020)Chandrasekaran, Feigenblat, Hovy,
  Ravichander, Shmueli-Scheuer, and
  de~Waard}]{chandrasekaran-etal-2020-overview-insights}
Muthu~Kumar Chandrasekaran, Guy Feigenblat, Eduard Hovy, Abhilasha Ravichander,
  Michal Shmueli-Scheuer, and Anita de~Waard. 2020.
\newblock \href {https://doi.org/10.18653/v1/2020.sdp-1.24} {Overview and
  insights from the shared tasks at scholarly document processing 2020:
  {CL}-{S}ci{S}umm, {L}ay{S}umm and {L}ong{S}umm}.
\newblock In \emph{Proceedings of the First Workshop on Scholarly Document
  Processing}, pages 214--224, Online. Association for Computational
  Linguistics.

\bibitem[{Cohan et~al.(2018)Cohan, Dernoncourt, Kim, Bui, Kim, Chang, and
  Goharian}]{cohan2018discourse}
Arman Cohan, Franck Dernoncourt, Doo~Soon Kim, Trung Bui, Seokhwan Kim, Walter
  Chang, and Nazli Goharian. 2018.
\newblock A discourse-aware attention model for abstractive summarization of
  long documents.
\newblock \emph{arXiv preprint arXiv:1804.05685}.

\bibitem[{Coleman and Liau(1975)}]{coleman1975computer}
Meri Coleman and Ta~Lin Liau. 1975.
\newblock A computer readability formula designed for machine scoring.
\newblock \emph{Journal of Applied Psychology}, 60(2):283.

\bibitem[{Crossley et~al.(2017)Crossley, Skalicky, Dascalu, McNamara, and
  Kyle}]{crossley2017predicting}
Scott~A Crossley, Stephen Skalicky, Mihai Dascalu, Danielle~S McNamara, and
  Kristopher Kyle. 2017.
\newblock Predicting text comprehension, processing, and familiarity in adult
  readers: New approaches to readability formulas.
\newblock \emph{Discourse Processes}, 54(5-6):340--359.

\bibitem[{Dathathri et~al.(2019)Dathathri, Madotto, Lan, Hung, Frank, Molino,
  Yosinski, and Liu}]{dathathri2019plug}
Sumanth Dathathri, Andrea Madotto, Janice Lan, Jane Hung, Eric Frank, Piero
  Molino, Jason Yosinski, and Rosanne Liu. 2019.
\newblock Plug and play language models: A simple approach to controlled text
  generation.
\newblock \emph{arXiv preprint arXiv:1912.02164}.

\bibitem[{Devaraj et~al.(2021)Devaraj, Marshall, Wallace, and
  Li}]{devaraj2021paragraph}
Ashwin Devaraj, Iain~J Marshall, Byron~C Wallace, and Junyi~Jessy Li. 2021.
\newblock Paragraph-level simplification of medical texts.
\newblock \emph{arXiv preprint arXiv:2104.05767}.

\bibitem[{Devlin et~al.(2019)Devlin, Chang, Lee, and
  Toutanova}]{DBLP:conf/naacl/DevlinCLT19}
Jacob Devlin, Ming{-}Wei Chang, Kenton Lee, and Kristina Toutanova. 2019.
\newblock \href {https://doi.org/10.18653/v1/n19-1423} {{BERT:} pre-training of
  deep bidirectional transformers for language understanding}.
\newblock In \emph{Proceedings of the 2019 Conference of the North American
  Chapter of the Association for Computational Linguistics: Human Language
  Technologies, {NAACL-HLT} 2019, Minneapolis, MN, USA, June 2-7, 2019, Volume
  1 (Long and Short Papers)}, pages 4171--4186. Association for Computational
  Linguistics.

\bibitem[{DeYoung et~al.(2021)DeYoung, Beltagy, van Zuylen, Kuehl, and
  Wang}]{deyoung2021ms2}
Jay DeYoung, Iz~Beltagy, Madeleine van Zuylen, Bailey Kuehl, and Lucy~Lu Wang.
  2021.
\newblock Ms2: Multi-document summarization of medical studies.
\newblock \emph{arXiv preprint arXiv:2104.06486}.

\bibitem[{Fan et~al.(2018)Fan, Grangier, and Auli}]{fan2018controllable}
Angela Fan, David Grangier, and Michael Auli. 2018.
\newblock Controllable abstractive summarization.
\newblock In \emph{Proceedings of the 2nd Workshop on Neural Machine
  Translation and Generation}, pages 45--54.

\bibitem[{Friedman et~al.(2002)Friedman, Kra, and Rzhetsky}]{friedman2002two}
Carol Friedman, Pauline Kra, and Andrey Rzhetsky. 2002.
\newblock Two biomedical sublanguages: a description based on the theories of
  zellig harris.
\newblock \emph{Journal of biomedical informatics}, 35(4):222--235.

\bibitem[{Gao et~al.(2019)Gao, Sun, and Passonneau}]{gao-etal-2019-automated}
Yanjun Gao, Chen Sun, and Rebecca~J. Passonneau. 2019.
\newblock \href {https://doi.org/10.18653/v1/K19-1038} {Automated pyramid
  summarization evaluation}.
\newblock In \emph{Proceedings of the 23rd Conference on Computational Natural
  Language Learning (CoNLL)}, pages 404--418, Hong Kong, China. Association for
  Computational Linguistics.

\bibitem[{Gehrmann et~al.(2019)Gehrmann, Strobelt, and Rush}]{gehrmann2019gltr}
Sebastian Gehrmann, Hendrik Strobelt, and Alexander~M Rush. 2019.
\newblock Gltr: Statistical detection and visualization of generated text.
\newblock In \emph{Proceedings of the 57th Annual Meeting of the Association
  for Computational Linguistics: System Demonstrations}, pages 111--116.

\bibitem[{Goyal et~al.(2021)Goyal, Rajani, Liu, and
  Kry{\'s}ci{\'n}ski}]{goyal2021hydrasum}
Tanya Goyal, Nazneen~Fatema Rajani, Wenhao Liu, and Wojciech
  Kry{\'s}ci{\'n}ski. 2021.
\newblock Hydrasum: Disentangling stylistic features in text summarization
  using multi-decoder models.
\newblock \emph{arXiv preprint arXiv:2110.04400}.

\bibitem[{Guo et~al.(2020)Guo, Qiu, Wang, and Cohen}]{guo2020automated}
Yue Guo, Wei Qiu, Yizhong Wang, and Trevor Cohen. 2020.
\newblock Automated lay language summarization of biomedical scientific
  reviews.
\newblock \emph{arXiv preprint arXiv:2012.12573}.

\bibitem[{He et~al.(2020)He, Kry{\'s}ci{\'n}ski, McCann, Rajani, and
  Xiong}]{he2020ctrlsum}
Junxian He, Wojciech Kry{\'s}ci{\'n}ski, Bryan McCann, Nazneen Rajani, and
  Caiming Xiong. 2020.
\newblock Ctrlsum: Towards generic controllable text summarization.
\newblock \emph{arXiv preprint arXiv:2012.04281}.

\bibitem[{Honnibal and Montani(2017)}]{spacy2}
Matthew Honnibal and Ines Montani. 2017.
\newblock {spaCy 2}: Natural language understanding with {B}loom embeddings,
  convolutional neural networks and incremental parsing.
\newblock To appear.

\bibitem[{Kincaid et~al.(1975)Kincaid, Fishburne~Jr, Rogers, and
  Chissom}]{kincaid1975derivation}
J~Peter Kincaid, Robert~P Fishburne~Jr, Richard~L Rogers, and Brad~S Chissom.
  1975.
\newblock Derivation of new readability formulas (automated readability index,
  fog count and flesch reading ease formula) for navy enlisted personnel.
\newblock Technical report, Naval Technical Training Command Millington TN
  Research Branch.

\bibitem[{Lewis et~al.(2019)Lewis, Liu, Goyal, Ghazvininejad, Mohamed, Levy,
  Stoyanov, and Zettlemoyer}]{lewis2019bart}
Mike Lewis, Yinhan Liu, Naman Goyal, Marjan Ghazvininejad, Abdelrahman Mohamed,
  Omer Levy, Ves Stoyanov, and Luke Zettlemoyer. 2019.
\newblock Bart: Denoising sequence-to-sequence pre-training for natural
  language generation, translation, and comprehension.
\newblock \emph{arXiv preprint arXiv:1910.13461}.

\bibitem[{Lewis et~al.(2020)Lewis, Liu, Goyal, Ghazvininejad, Mohamed, Levy,
  Stoyanov, and Zettlemoyer}]{DBLP:conf/acl/LewisLGGMLSZ20}
Mike Lewis, Yinhan Liu, Naman Goyal, Marjan Ghazvininejad, Abdelrahman Mohamed,
  Omer Levy, Veselin Stoyanov, and Luke Zettlemoyer. 2020.
\newblock \href {https://doi.org/10.18653/v1/2020.acl-main.703} {{BART:}
  denoising sequence-to-sequence pre-training for natural language generation,
  translation, and comprehension}.
\newblock In \emph{Proceedings of the 58th Annual Meeting of the Association
  for Computational Linguistics, {ACL} 2020, Online, July 5-10, 2020}, pages
  7871--7880. Association for Computational Linguistics.

\bibitem[{Lin(2004)}]{lin2004rouge}
Chin-Yew Lin. 2004.
\newblock Rouge: A package for automatic evaluation of summaries.
\newblock In \emph{Text summarization branches out}, pages 74--81.

\bibitem[{Liu and Lapata(2019{\natexlab{a}})}]{DBLP:conf/emnlp/LiuL19}
Yang Liu and Mirella Lapata. 2019{\natexlab{a}}.
\newblock \href {https://doi.org/10.18653/v1/D19-1387} {Text summarization with
  pretrained encoders}.
\newblock In \emph{Proceedings of the 2019 Conference on Empirical Methods in
  Natural Language Processing and the 9th International Joint Conference on
  Natural Language Processing, {EMNLP-IJCNLP} 2019, Hong Kong, China, November
  3-7, 2019}, pages 3728--3738. Association for Computational Linguistics.

\bibitem[{Liu and Lapata(2019{\natexlab{b}})}]{liu2019text}
Yang Liu and Mirella Lapata. 2019{\natexlab{b}}.
\newblock Text summarization with pretrained encoders.
\newblock In \emph{Proceedings of the 2019 Conference on Empirical Methods in
  Natural Language Processing and the 9th International Joint Conference on
  Natural Language Processing (EMNLP-IJCNLP)}, pages 3730--3740.

\bibitem[{Martinc et~al.(2021{\natexlab{a}})Martinc, Pollak, and
  Robnik-{\v{S}}ikonja}]{martinc2021supervised}
Matej Martinc, Senja Pollak, and Marko Robnik-{\v{S}}ikonja.
  2021{\natexlab{a}}.
\newblock Supervised and unsupervised neural approaches to text readability.
\newblock \emph{Computational Linguistics}, 47(1):141--179.

\bibitem[{Martinc et~al.(2021{\natexlab{b}})Martinc, Pollak, and
  Robnik-Šikonja}]{Martinc2021}
Matej Martinc, Senja Pollak, and Marko Robnik-Šikonja. 2021{\natexlab{b}}.
\newblock \href {https://doi.org/10.1162/coli_a_00398} {Supervised and
  unsupervised neural approaches to text readability}.
\newblock \emph{Computational Linguistics}, 47:141--179.

\bibitem[{Maynez et~al.(2020)Maynez, Narayan, Bohnet, and
  McDonald}]{maynez2020faithfulness}
Joshua Maynez, Shashi Narayan, Bernd Bohnet, and Ryan McDonald. 2020.
\newblock On faithfulness and factuality in abstractive summarization.
\newblock \emph{arXiv preprint arXiv:2005.00661}.

\bibitem[{Plavén-Sigray et~al.(2017)Plavén-Sigray, Matheson, Schiffler, and
  Thompson}]{10.7554/eLife.27725}
Pontus Plavén-Sigray, Granville~James Matheson, Björn~Christian Schiffler,
  and William~Hedley Thompson. 2017.
\newblock \href {https://doi.org/10.7554/eLife.27725} {Research: The
  readability of scientific texts is decreasing over time}.
\newblock \emph{eLife}, 6:e27725.

\bibitem[{Senter and Smith(1967)}]{senter1967automated}
RJ~Senter and Edgar~A Smith. 1967.
\newblock Automated readability index.
\newblock Technical report, Cincinnati Univ OH.

\bibitem[{Sotudeh et~al.(2020)Sotudeh, Goharian, and
  Filice}]{sotudeh2020attend}
Sajad Sotudeh, Nazli Goharian, and Ross~W Filice. 2020.
\newblock Attend to medical ontologies: Content selection for clinical
  abstractive summarization.
\newblock \emph{arXiv preprint arXiv:2005.00163}.

\bibitem[{Wallace et~al.(2020)Wallace, Saha, Soboczenski, and
  Marshall}]{DBLP:journals/corr/abs-2008-11293}
Byron~C. Wallace, Sayantan Saha, Frank Soboczenski, and Iain~James Marshall.
  2020.
\newblock \href {http://arxiv.org/abs/2008.11293} {Generating (factual?)
  narrative summaries of rcts: Experiments with neural multi-document
  summarization}.
\newblock \emph{CoRR}, abs/2008.11293.

\bibitem[{Wang et~al.(2020{\natexlab{a}})Wang, Cho, and Lewis}]{wang2020asking}
Alex Wang, Kyunghyun Cho, and Mike Lewis. 2020{\natexlab{a}}.
\newblock Asking and answering questions to evaluate the factual consistency of
  summaries.
\newblock In \emph{Proceedings of the 58th Annual Meeting of the Association
  for Computational Linguistics}, pages 5008--5020.

\bibitem[{Wang et~al.(2021)Wang, Xie, Pei, Tiwari, Li et~al.}]{wang2021pre}
Benyou Wang, Qianqian Xie, Jiahuan Pei, Prayag Tiwari, Zhao Li, et~al. 2021.
\newblock Pre-trained language models in biomedical domain: A systematic
  survey.
\newblock \emph{arXiv preprint arXiv:2110.05006}.

\bibitem[{Wang et~al.(2020{\natexlab{b}})Wang, Lo, Chandrasekhar, Reas, Yang,
  Eide, Funk, Kinney, Liu, Merrill et~al.}]{wang2020cord}
Lucy~Lu Wang, Kyle Lo, Yoganand Chandrasekhar, Russell Reas, Jiangjiang Yang,
  Darrin Eide, Kathryn Funk, Rodney Kinney, Ziyang Liu, William Merrill, et~al.
  2020{\natexlab{b}}.
\newblock Cord-19: The covid-19 open research dataset.
\newblock \emph{ArXiv}.

\bibitem[{Xie et~al.(2022{\natexlab{a}})Xie, Bishop, Tiwari, and
  Ananiadou}]{xie2022pre}
Qianqian Xie, Jennifer~Amy Bishop, Prayag Tiwari, and Sophia Ananiadou.
  2022{\natexlab{a}}.
\newblock Pre-trained language models with domain knowledge for biomedical
  extractive summarization.
\newblock \emph{Knowledge-Based Systems}, 252:109460.

\bibitem[{Xie et~al.(2022{\natexlab{b}})Xie, Huang, Saha, and
  Ananiadou}]{xie2022gretel}
Qianqian Xie, Jimin Huang, Tulika Saha, and Sophia Ananiadou.
  2022{\natexlab{b}}.
\newblock Gretel: Graph contrastive topic enhanced language model for long
  document extractive summarization.
\newblock \emph{arXiv preprint arXiv:2208.09982}.

\bibitem[{Zhang et~al.(2019)Zhang, Gong, Yan, Duan, Xu, Wang, Gong, and
  Zhou}]{DBLP:journals/corr/abs-1902-09243}
Haoyu Zhang, Yeyun Gong, Yu~Yan, Nan Duan, Jianjun Xu, Ji~Wang, Ming Gong, and
  Ming Zhou. 2019.
\newblock \href {http://arxiv.org/abs/1902.09243} {Pretraining-based natural
  language generation for text summarization}.
\newblock \emph{CoRR}, abs/1902.09243.

\bibitem[{Zheng et~al.(2020)Zheng, Cai, Zhang, and
  Li}]{DBLP:conf/coling/ZhengCZL20}
Changmeng Zheng, Yi~Cai, Guanjie Zhang, and Qing Li. 2020.
\newblock \href {https://doi.org/10.18653/v1/2020.coling-main.497}
  {Controllable abstractive sentence summarization with guiding entities}.
\newblock In \emph{Proceedings of the 28th International Conference on
  Computational Linguistics, {COLING} 2020, Barcelona, Spain (Online), December
  8-13, 2020}, pages 5668--5678. International Committee on Computational
  Linguistics.

\end{thebibliography}

\clearpage
\appendix
\section{Experiment Setup}
\label{sec:es}
All our experiments were run at a single NVIDIA Tesla A-100 GPU. 
We set input length as 8,192 for covering the whole text of our source document in LED and chose a learning rate of $3e-5$ with warm-up for 3 epochs to finetune all the models from the checkpoint in hugging face \footnote{\url{https://huggingface.co/allenai/led-base-16384}}.
 With regard to the optimizer, we used AdamW. For generation, we use a beam size of 4, with no repeat n-gram equal to 3. In extractive methods, we select top\textit{k} sentences when the candidate summary reaches the average token number of technical summaries and PLS. All our models are built with PyTorch and HuggingFace.

\section{RNPTC Algorithm}
\label{sec:al}
\begin{algorithm}
\caption{To compute a text complexity score given a document $d$ and a PLM $lm$. The FORWARD function returns the output matrix where each row maps to a distribution over all the tokens in the vocabulary. The APPEND function adds a value in the list. The RANKMEAN function is calculated as in Equation \ref{eq2}.}\label{alg:cap}
\small
\begin{algorithmic}[1]
\Procedure{RNPTC}{$d, lm$}
\State {NPs $\gets $ Noun phrases list extracted from d}
\State {P $\gets $ Create empty NP probability list}
\For{$i = \{1,\cdots,|NPs|\}$}
    \State {$T \gets $ Token sequence of NPs[i]}
    \State {$d' \gets d$}
    \State {$p \gets $ Create empty token probability list}
    \ForAll{$t \in T$}
        \State {$d'[t] \gets$ [MASK]}
    \EndFor
    \State {$output \gets$ FORWARD$(lm, d'$)}
    \ForAll{$t \in T$}
        \State{$prob \gets output[t][d[t]]$}
        \State{APPEND($p, prob$)}
    \EndFor
    \State {APPEND($P$, mean($p$))}
\EndFor
\State{$\hat{P} \gets$ descending sort($P$)}
\State{\textbf{return}  - Log(RANKMEAN($\hat{P}$))}
\EndProcedure
\end{algorithmic}
\end{algorithm}

\section{CDSR Collecting}
\label{sec append cdsr clt}
The Cochrane Database of Systematic Reviews (CDSR) includes peer-reviewed systematic reviews covering various topics in the healthcare domain. In particular, each review in CDSR contains an abstract and a plain language summary which has been required from authors
submitting a review since 2015. Prior to this, they were written by Cochrane staff with specialized training.
Following the collecting process introduced by \citet{guo2020automated}\footnote{https://github.com/qiuweipku/Plain
language summarisation} we extracted 8442 abstracts paired with their PLS from CDSR reviews from up to 14th Sept 2022. The data can be downloaded from the official API\footnote{https://www.cochranelibrary.com/cdsr/reviews} but it may differ from our experimented reviews due to the change from Cochrane.
The average length of our collected CDSR abstracts by word is around 721 while the average length of PLS is about 395.

\section{Human Evaluation Setup}
\label{sec:appdx human}
The evaluation samples are 10 documents drawn from the test set with their pairs of target summaries and the corresponding pairs of generated summaries by the multi-heads abstractive model (MH Abs) due to its highest ROUGE scores and lowest n-grams overlapping. 

Our standard for eligible evaluators is being able to read and write in academic English and having an undergraduate degree. Specifically, we recruit four \textit{experts} who have a medical undergraduate degree and either have or pursue a doctorate in the biomedical area and another four \textit{laypeople} with no biomedical related degree or experience in medical institutions. The final score of each aspect is averaged across tested documents and marking evaluators.

\clearpage
\onecolumn
\section{Example Output}
\label{sec:appdx examples}
\begin{table*}[hbt!]
\small
\begin{tabular}{m{0.9\textwidth}}
\hline \textbf{Target Technical Summary} Sex differences in schizophrenia are well known, but their genetic basis has not been identified. We performed a genome-wide association scan for schizophrenia in an Ashkenazi Jewish population using DNA pooling. We found a female-specific association with rs7341475, a SNP in the fourth intron of the reelin (RELN) gene (p = 2.9 × 10−5 in women), with a significant gene-sex effect (p = 1.8 × 10−4). We studied rs7341475 in four additional populations, totaling 2,274 cases and 4,401 controls. A significant effect was observed only in women, replicating the initial result (p = 2.1 × 10−3 in women; p = 4.2 × 10−3 for gene-sex interaction). Based on all populations the estimated relative risk of women carrying the common genotype is 1.58 (p = 8.8 × 10−7; p = 1.6 × 10−5 for gene-sex interaction). The female-specific association between RELN and schizophrenia is one of the few examples of a replicated sex-specific genetic association in any disease.
\\\hline
\textbf{Generated Technical Summary} Schizophrenia is a common psychiatric disorder of unknown aetiology. Sex differences in the risk of developing schizophrenia have been reported for a number of loci, but the robustness of these claims is open to doubt. We have carried out a genome-wide association study of schizophrenia using DNA pools of cases and controls from the Ashkenazi Jewish population. We identified a single SNP in Chromosome 7 (rs7341475) in the fourth intron of the reelin gene (RELN) that was replicated in the UK case-control sample with an effect in the same direction (p = 1.3×10−4). The odds ratio (ORG) for women in the combined replication set was 1.41 (95\% confidence interval [CI] 1.3–1.6) for men and ORG of 0.97 for women. We tested the association in four other sample sets, three of European ancestry and one of Chinese descent. We found no evidence for stratification in the US sample, indicating that the risk for schizophrenia cannot be caused by population structure. Our result of a female-specific association of RELN with schizophrenia may suggest a possible pathway where sex hormones modulate gene expression, which by altering cortical structure, increases susceptibility to psychosis.
\\\hline
\textbf{Target Plain Language Summary} Schizophrenia is a complex mental disease, which includes symptoms of delusions, hallucinations, disorganized speech, aberrant behavior, lack of emotional expression, diminished motivation, and social withdrawal. The cause of schizophrenia is unknown, but there is extensive evidence that genetics play a significant role in its aetiology. We studied the genetic basis of schizophrenia by analyzing around 500,000 genetic variants distributed across the whole human genome in DNA from schizophrenic patients and controls. We analyzed separately the DNA from men and women, and identified a genetic variant that increases the risk of developing schizophrenia in women only. The genetic variant is estimated to increase the risk of schizophrenia for women carrying the risk variant by 1.4-fold. The genetic variant is in a gene called reelin, which is known to play a part in brain development. However, it is still unclear how this genetic variant predisposes to schizophrenia nor why it is specific to women only.\\\hline
\textbf{Generated Plain Language Summary} Schizophrenia is a common psychiatric disorder of unknown aetiology. The heritability of schizophrenia is approximately 80\%. However, sex differences in the risk of developing schizophrenia have so far been reported for a number of loci, but the robustness of these claims is open to doubt; results have yet to be corroborated. In this study, we carried out a genome-wide association study using DNA pools of cases and controls from the Ashkenazi Jewish population in Northern Ireland and the Republic of Ireland. We identified one SNP in the reelin gene, rs7341475, in the fourth intron of the gene. The SNP is located in Chromosome 7 (RELN), a gene previously studied for association with schizophrenia. We found that the frequency of this SNP is increased in women with schizophrenia, suggesting a possible pathway where sex hormones modulate gene expression, which by altering cortical structure, increases susceptibility to psychosis. Our result of a female-specific association of RELN with schizophrenia may suggest that sex hormones may mediate changes in the function of the RELN gene.
\\\hline
\end{tabular}
\label{table: examples_}
\end{table*}

\begin{table*}[hbt!]
\small
\begin{tabular}{m{0.9\textwidth}}
\hline \textbf{Target Technical Summary} Pathogen perception by the plant innate immune system is of central importance to plant survival and productivity. The Arabidopsis protein RIN4 is a negative regulator of plant immunity. In order to identify additional proteins involved in RIN4-mediated immune signal transduction, we purified components of the RIN4 protein complex. We identified six novel proteins that had not previously been implicated in RIN4 signaling, including the plasma membrane (PM) H+-ATPases AHA1 and/or AHA2. RIN4 interacts with AHA1 and AHA2 both in vitro and in vivo. RIN4 overexpression and knockout lines exhibit differential PM H+-ATPase activity. PM H+-ATPase activation induces stomatal opening, enabling bacteria to gain entry into the plant leaf; inactivation induces stomatal closure thus restricting bacterial invasion. The rin4 knockout line exhibited reduced PM H+-ATPase activity and, importantly, its stomata could not be re-opened by virulent Pseudomonas syringae. We also demonstrate that RIN4 is expressed in guard cells, highlighting the importance of this cell type in innate immunity. These results indicate that the Arabidopsis protein RIN4 functions with the PM H+-ATPase to regulate stomatal apertures, inhibiting the entry of bacterial pathogens into the plant leaf during infection.

\\\hline
\textbf{Generated Technical Summary} Plant innate immunity is essential for the survival of many pathogens. However, how pathogen-associated molecular patterns (PAMPs) and effector-triggered immunity (ETI) interact remains elusive. Here, we report the identification and characterization of the Arabidopsis RIN4 protein complex in planta. We demonstrate that the rin4 knockout line exhibits decreased PM H+-ATPase activity, whereas the rins2/rin4 double mutant line exhibits increased AHA1 and AHA2 activity. We also show that RIN1/AHA2 is expressed in guard cells along with other PTI and ETI signaling components. Our results support a model in which RIN2 associates with the C-terminal autoinhibitory domain of the PMH+-atPase to regulate leaf stomatal apertures in response to pathogen effectors.
\\\hline
\textbf{Target Plain Language Summary} Plants are continuously exposed to microorganisms. In order to resist infection, plants rely on their innate immune system to inhibit both pathogen entry and multiplication. We investigated the function of the Arabidopsis protein RIN4, which acts as a negative regulator of plant innate immunity. We biochemically identified six novel RIN4-associated proteins and characterized the association between RIN4 and the plasma membrane H+-ATPase pump. Our results indicate that RIN4 functions in concert with this pump to regulate leaf stomata during the innate immune response, when stomata close to block the entry of bacterial pathogens into the leaf interior.
\\\hline
\textbf{Generated Plain Language Summary} Plants are continuously exposed to a variety of microorganisms. In order to successfully avoid infection, they have evolved a series of defense mechanisms that work in concert to limit pathogen invasion and multiplication. The most well-studied bacterial effectors come from P. syringae pv. tomato (Pst), the causal agent of bacterial speck on Arabidopsis thaliana. Pst utilizes the type III secretion system (PTI) to deliver effector proteins into the plant cell during infection, resulting in effector-triggered immunity (ETI). However, how pathogen perception activates immune responses and signaling overlap between PTI and ETI remains elusive. In this study, we report the identification and characterization of the RIN4 protein complex in planta. We identified the PM H+-ATPases AHA1 and AHA2, whose interaction we characterized in greater detail. We also demonstrate that the rin4 knockout line cannot re-open its stomata in response to virulent Pst. Importantly, we also show that Rin4 is expressed in guard cells along with other PTI signaling components. Our findings are consistent with a model in which RIN 4 associates with the C-terminal autoinhibitory domain (AHA1/AHA2) to regulate leaf stomatal apertures in
\\\hline
\end{tabular}
\caption{Examples of target summary and generated output from Mutli Heads Abstractive Model}
\label{table: examples}
\end{table*}
\end{document}